# ToolEENet: Tool Affordance 6D Pose Estimation

Yunlong Wang[1†], Lei Zhang[1,2†], Yuyang Tu[1*], Hui Zhang[1], Kaixin Bai[1,2], Zhaopeng Chen[2] and Jianwei Zhang[1]

*Abstract*— The exploration of robotic dexterous hands utilizing tools has recently attracted considerable attention. A significant challenge in this field is the precise awareness of a tool's pose when grasped, as occlusion by the hand often degrades the quality of the estimation. Additionally, the tool's overall pose often fails to accurately represent the contact interaction, thereby limiting the effectiveness of vision-guided, contact-dependent activities. To overcome this limitation, we present the innovative TOOLEE dataset, which, to the best of our knowledge, is the first to feature affordance segmentation of a tool's end-effector (EE) along with its defined 6D pose based on its usage. Furthermore, we propose the ToolEENet framework for accurate 6D pose estimation of the tool's EE. This framework begins by segmenting the tool's EE from raw RGB-D data, then uses a diffusion model-based pose estimator for 6D pose estimation at a category-specific level. Addressing the issue of symmetry in pose estimation, we introduce a symmetry-aware pose representation that enhances the consistency of pose estimation. Our approach excels in this field, demonstrating high levels of precision and generalization. Furthermore, it shows great promise for application in contact-based manipulation scenarios. All data and codes are available on the project website: https://tooleenet-iros2024.github.io/

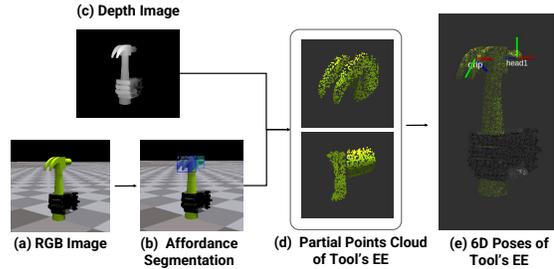

Fig. 1. We present a framework for estimating the 6D pose of the tool's end-effectors (EE). The framework first estimates the (b) from (a) by segmentation algorithm, then it combines (c) to extract the (d) partial points cloud of the tool's EE. Finally, (e) is obtained by a category-level pose estimation based on the diffusion model from the points cloud.

## I. INTRODUCTION

Object state estimation is crucial for both simulated and real-world robotic manipulation. This process encompasses identifying the object's 2D/3D bounding box and its 6D pose, among other factors. The 6D pose, which includes both translation and rotation information, is especially vital to robotic manipulation. Previous research on 6D pose estimation has primarily focused on instance-level [1]–[3] and category-level [4]–[6] estimations, employing an object-centric approach as standard. This approach has been adequate for fundamental manipulation techniques [7].

In contact-based manipulation tasks, such as a robot using a hammer to drive a nail, the interaction between the hammerhead and the nail is crucial. Thus, accurately determining the hammerhead's pose becomes a fundamental component of the task. The hammerhead, analogous to a robot arm's end-effector (EE), is integral in generating direct contact effects during manipulation. A common approach to estimating the EE's pose is through instance-level pose estimation to determine the object's 6D pose. Due to the object-centric nature of the pose, additional priors are necessary to deduce the transformation from the object's pose to its EE. The need for individual pose priors for each 3D model instance limits the generalization of learned skills. To address this, category-level pose estimation has been developed. This method can estimate poses within a specific category, but it still yields an object-centric pose, and the manipulation task has been limited [8]. The variability in shape and scale within a category impedes the effective application of category-level priors for determining the EE's pose.

Our conceptualization posits that despite the significant variability in shape and scale of objects at the category level, the geometry of the tool's end-effector (EE) exhibits a higher degree of consistency. For instance, the hammerhead typically maintains a roughly cylindrical shape. Predicated on this assumption, we hypothesize that category-level pose estimation, when fed with the tool's EE point cloud as input, could achieve remarkable accuracy in determining EE's pose.

In pursuit of this objective, we have built a synthetic dataset specifically designed for this pose estimation task. The dataset includes poses of the tool's EE that correlate with its contact interactions. Additionally, we have developed a framework for estimating the pose of the tool's EE based on its affordance. As seen in Fig. 1, this process begins with the prediction of affordance segmentation from RGB images using the Mask R-CNN algorithm [9]. This segmentation, in combination with the depth image, is used to extract a partial point cloud of each tool's EE. Subsequently, a generative diffusion model is employed to estimate the 6D pose of the tool's EE. To address the challenges caused by symmetry in pose estimation, we introduce a symmetry-aware rotational representation. This approach involves estimating

[1] Department of Informatics, Universität Hamburg, Hamburg 20146, Germany.
[2] Agile Robots AG.
† The first two authors contributed equally to this work.
* Yuyang Tu is the corresponding author. Email: tu@informatik.uni-hamburg.de

This research was funded by the German Research Foundation (DFG) and the National Science Foundation of China (NSFC) in project Crossmodal Learning, DFG TRR-169/NSFC.

the symmetry properties ($S_x$,$S_y$,$S_z$) alongside the rotations ($R_x$,$R_y$,$R_z$), enabling the algorithm to autonomously determine the symmetrical axis for metric calculation and pose clarification. Our main contributions can be summarized as follows:

- We develop a novel synthetic dataset specifically tailored for affordance pose estimation, particularly focusing on applications involving dexterous tool manipulation. This dataset facilitates an in-depth examination of the influence of geometric variability, especially in the context of point cloud data.
- We propose an innovative framework adept at accurately estimating the 6D pose of the tool's end-effector (EE) from RGB-D inputs. This framework incorporates our newly proposed symmetry-aware pose representation, effectively addressing and resolving issues related to symmetry in pose estimation.
- We conduct comprehensive experiments to validate the robustness and effectiveness of our methodology in estimating the pose of the tool's EE. These experiments demonstrate the high reliability and superior performance of our approach, especially in symmetric cases.

## II. RELATED WORK

### A. Category Level Object Pose Estimation

Unlike most instance-level pose estimation, category-level pose estimation does not require an exact 3D model of the object. However, it is particularly challenging due to the wide variety of shapes, scales, and appearances within the same category. To this end, NOCS [4] first introduces a novel representation that maps objects of various categories to a normalized space, facilitating the estimation of their pose and size across different instances. Some pose regression and shape reconstruction methods [10], [11] have been used for the improvement regarding shape variance, [5], [12] use the key-points matching strategy to resolve the category-level pose estimation problem, and [13] leverages both geometric and semantic features from pre-trained models to enhance pose estimation accuracy with minimal training data. However, the direct regression method usually only works well with small intra-class variance. Generative models are more promising to handle these problems, to address the multi-hypothesis issue prevalent in partially observed point clouds and symmetric objects in category-level pose estimation, GenPose [6] introduces a generative model approach using diffusion models and reformulates the pose estimation to a conditional generation question. Moreover, [14] uses the pose-aware generative model to do analysis-by-synthesis tasks to pair the generated image with different poses that best agree with the observation. For category-level object pose estimation, another challenge is to generate high-quality training data, because the training datasets should include a wide range of variations in categories, instances, poses, clutters, and domain conditions. Real-world datasets [15], [16], often fall short in terms of size and diversity due to the prohibitive costs of manual annotation. Furthermore, collecting data in real-world tool manipulation scenarios by robot hand is expensive [17]. Consequently, synthetic data generation is an alternative. To bridge the domain gap with real-world data, some efforts, such as [4], [18], have mixed real and synthetic data by rendering virtual objects onto real backgrounds. In our work, we synthesize the high-fidelity dataset by using Isaac Gym [19] and domain randomization to simulate the robot hand-grasping tool in the context of tool manipulation, which enables the creation of large, realistic, well-annotated datasets with low cost.

### B. Object Affordances Detection

Object affordance detection aims to discern the potential actions that objects can support. This concept, deeply rooted in ecological psychology, has been adapted to enhance the interaction capabilities of robots and intelligent systems with their environments [20]. By recognizing the affordances of objects, these systems can make informed decisions on how to manipulate or utilize them across various scenarios. Some previous work [21], [22] use Convolutional Neural Networks (CNNs) based approaches for direct affordance detection from RGB images. Human-centric methods [23], [24] focus on learning from human interactions, suggesting that observing how humans use objects can provide critical insights into their affordances. Furthermore, as discussed by [25], simulation-based learning environments offer new avenues for affordance detection by allowing models to learn from virtual interactions. Previous works have integrated affordance detection with grasping, where grasp frames are generated alongside affordance detection to make the robot aware of the object it grasps [22], [26], [27]. In contrast, our research underscores the importance of combining tool affordances with their 6D poses. This approach not only enables robots to understand the function of a tool but also its precise spatial position and orientation, which is crucial for its effective application. Such integration significantly enhances robotic dexterity and accuracy in tool manipulation.

## III. TOOLEE DATASET

Many datasets [4], [28], [29] are collected to estimate the object-centric pose, and algorithms based on those datasets are insufficient for assisting the contact-based task at category-level. To this end, we proposed the TOOLEE dataset, a high-fidelity dataset for estimating the 6D pose from the RGB-D input. The dataset contains the segmentation of each tool's EE, the depth image, the RGB image, the hand-occluded partial points cloud, the corresponding annotation for both the tool and the tool's EE, also some descriptive information.

### A. Dataset Setup

**For Gathering the Standard 3D Model Assets:** We downloaded many 3D printable mesh models from the open-source database and applied diagonal normalization and zero means to them. For the orientation of each asset, we manually adjust it to have a consistent pose for objects within a category.

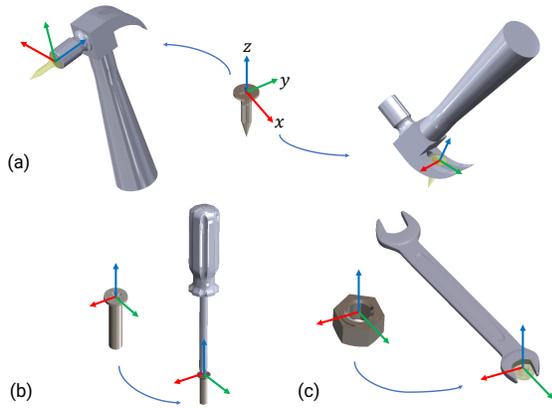
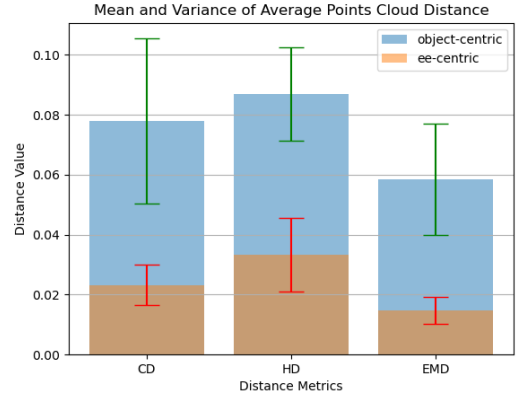

Fig. 2. The annotation of coordinate frame of the Tool's EE by semantic of the tool usage. (a) Shows the coordinate frame of the hammer's EE defined by the exact same coordinate frame of the nail when the hammer has interacted (hammering or pulling) with the nail. (b) (c) Shows the definition of the coordinate frame of screw driver's EE and wrench's EE.

Fig. 3. The overall distance measurement of object-centric and EE-centric points cloud. CD denotes Chamfer distance, HD denotes Hausdorff distance and the EMD denotes the Earth-Mover's distance. The blue bar and green line denote the mean and variance distance of the object-centric points cloud, and the orange bar and red line denote the mean and variance of the EE-centric points cloud.

**For 6D Pose Annotation:** Fig. 2. shows the definition of the coordinate frame of each tool's EE. Since we focus on the contact paradigm between the tool and the object, we choose to use the object pose at the contact moment as the tool's EE pose. And we manually annotate EE's 6D pose of each CAD model. Moreover, affordance segmentation is generated by drawing a circle around the EE point projected in an RGB image and taking the intersection with tool's segmentation.

**For Rendering the Dataset:** We roughly fixed the object and robot hand at a certain pose and randomized the position of the camera to gather RGB-D and segmentation images. For affordance segmentation, we simply chose the circle area on the RGB image around the end-effector point mapped from the 3D world. Even this could incorporate not only the end-effector of the tool but, in practice, it can potentially reduce the symmetry of the points cloud, and lead to a better rotation estimation. Furthermore, to reduce the bias of the dataset, we randomly translated the object along the z-axis to simulate the different grasp poses and occlusion effects. We also adjusted the object's scale within a predetermined range to ensure diversity in its appearance.

**For Dataset Cleaning:** Due to the variety of the camera view, some end-effectors of the tool could be occluded, we count the extracted points of each end-effector and filter out the example if less than 50 points. And we augment the points cloud to 1024 points by resampling it.

### B. Dataset Inspection

In this section, we first briefly introduce the distribution of the dataset, then try to go through the quantitative analysis to verify our assumption and to support the rationality of our proposed method.

*1) Dataset Distribution:* The synthetic dataset contains 4 categories of tool's EE, including the head of the hammer, grip of the hammer, head of the screwdriver, and head of the wrench. Each category contains 10 to 20 instance-level 3D Models. The whole dataset contains roughly in total 58000 examples.

*2) Dataset Analysis:* To test our method's underlying assumption that within a category, the shapes and sizes of objects exhibit significant variance, which will lead to imprecise estimation of the tool's EE pose, but the shapes and scales of the tool's EE are more consistent, allowing for more accurate pose predictions. Since the comparison of shapes and scales between the object and the tool' EE can not directly reflect the impact of the pose estimation, we first extract the tool's EE points cloud and convert them into two classes, object-centric and EE-centric. In this manner, the distance between the points cloud then can reflect the influence from the scale and shape variety of objects and tool's EE at a certain level.

For evaluating the distances between points clouds, we use the following three metrics:

- **Chamfer Distance(CD)**: It measures the average distance between the pairs of Nearest Neighbours between $P_1 = \{x_i \in \mathbb{R}^3\}_{i=1}^n$ and $P_2 = \{x_j \in \mathbb{R}^3\}_{j=1}^m$.
- **Hausdorff Distance(HD)**: It measures the maximum distance between any pair of Nearest Neighbours between $P_1$ and $P_2$.
- **Earth-Mover's Distance(EMD)**: It measures the average distance between pairs of points between $P_1$ and $P_2$ according to an optimal correspondence $\pi \in \prod(P, Q)$, which is the $n \times m$ matrices where the rows and columns sum to one.

Fig. 3. illustrates that in terms of both mean and variance, the EE-centric point cloud demonstrates greater consistency compared to the object-centric point cloud. Based on these findings, we can affirm that our hypothesis is tenable. In the subsequent sections, we will investigate whether this hypothesis allows for accurate estimation of the tool's EE poses, specifically by utilizing partial point clouds of the EE as input, instead of the entire object's point cloud.

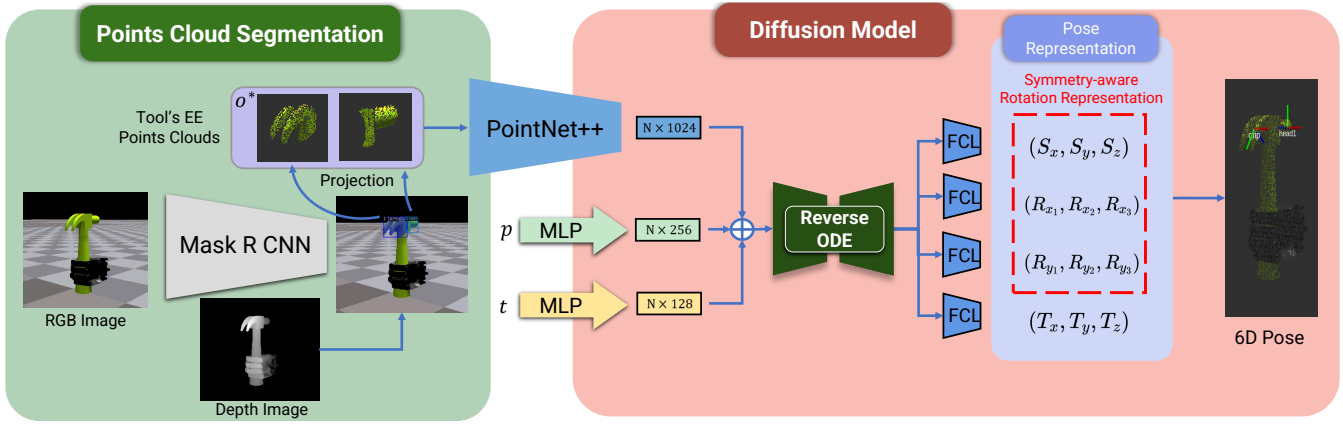

Fig. 4. The whole pipeline of ToolEENet. We first employ the Mask R-CNN to segment the part of the tool's EE from the RGB input. Then by the projection of the depth input, the partial points clouds of the tool's EE can be acquired. The points cloud $O$ are processed by PointNet++ to extract the feature, then combined with the input pose $p$ and time variable $t$, to generate the candidate pose using a defusion modal. The pose is in the symmetric agnostic rotation representation as we introduced.

## IV. METHOD

In this research, our goal is to estimate the 6D pose of a tool's end effector (EE) from the observed RGB-D image. Firstly we use the RGB image to get semantic segmentation of the tool and the tool's EE from the in-hand scenario. Using the segmentation and depth image to project the observed point cloud $\mathbf{O}$ to get each tool's EE point cloud $\bigcup_{j=1}^{k} \mathbf{O}_j \subset \mathbf{O}$, here $k$ represent to the number of the end effector for each tool, for example in hammer category $k = 2$. Then we had a trained model given the paired tool's EE pose and its observed point cloud $D = \{(p_{ij}, O_j)\}_{i=1, j=1}^{n,k}$, where $p_j \in$ SE(3) and $O_j \in \mathbb{R}^{3 \times N_j}$ denote 6D pose of the $j$th segmented 3D point cloud of tool's EE with $N_j$ points respectively. Given an unseen point cloud of robot tool usage $O^*$, our goal is to estimate the tool's EE ground truth pose $p_j^*$. Fig. 4. overviews the proposed pipeline.

### A. Tool's EE Segmentation

For the affordance segmentation task, we simply utilized the **Mask R-CNN** [9]. Given an RGB-D image, we first input the RGB component into Mask R-CNN to obtain the semantic segmentation, i.e., end effectors of this tool. Subsequently, we use semantic segmentation to project the depth image to obtain the points cloud of each tool's EE. Each point cloud is sampled to 1028 points and then fed into the 6D pose estimation model. Since the features of the tool's semantic parts are easily captured, this approach can even be generalized to previously unseen objects at a categorical level.

### B. Diffusion Model

We use a diffusion model to generate multiple candidate poses and use mean pooling to get the estimation of the pose from an unseen partial point cloud $O^*$. we assume our ground truth data is sampled from a joint distribution of a pose and corresponding point clouds $D = \{(p_i, O_i) \sim P(p, O)\}$. Our

**Algorithm 1** Pose Estimation with Diffusion Model

1: **Training Phase:**
2: Initialize score network $\Phi_\theta$ with parameters $\theta$.
3: Define SDE parameters: $\sigma_{\min}$, $\sigma_{\max}$, and noise level $\epsilon$.
4: **for** each training iteration **do**
5:     Sample a batch of point clouds $\{O_i\}$ and corresponding ground truth poses $\{p(0)_i\}$.
6:     Sample $t \sim \mathcal{U}(\epsilon, 1)$.
7:     Perturb ground truth poses to obtain $\{p(t)_i\}$ using the SDE.
8:     Compute loss $L(\theta)$ according to Eq. (1).
9:     Update $\Phi_\theta$ to minimize $L(\theta)$.
10: **end for**
11:
12: **Sampling Phase:**
13: **for** each unseen point cloud $O^*$ **do**
14:     Initialize pose $p(1)$ from $\mathcal{N}(0, \sigma_{\max}^2 I)$.
15:     Solve the ODE from $t = 1$ to $t = \epsilon$ to update $p(t)$ according to Eq. (2).
16:     Estimate Pose candidates $\{\hat{p}_i\}_{i=1}^{K} \sim P(p, O)$
17:     Mean Pooling $\bar{p} \leftarrow \frac{1}{K} \sum_{i=1}^{K} \hat{p}_i$
18: **end for**
19:
20: **Return:** Estimated poses $\hat{p}$ for unseen point clouds.

task is to model the conditional pose distribution $P(p \mid O)$ during the training. For the inference stage, we are using that distribution to sample the estimated pose from that unseen point cloud $P(p \mid O^*)$. In practice, we are using a score-based diffusion model to model conditional distribution. Algorithm 1 shows the whole process. In detail, we use a Stochastic Differential Equation (SDE) proposed by [30] to generate perturbed poses. We use the objective proposed by [6] as the loss for the training.

$$\mathcal{L}(\theta) = \mathbb{E}_{t \sim \mathcal{U}(\epsilon,1)} \left\{ \lambda(t) \mathbb{E}_{\substack{p(0) \sim P(p(0)|O), \\ p(t) \sim \mathcal{N}(p(t); p(0), \sigma^2(t)I)}} \left[ \left\| \Phi_\theta(p(t), t \mid O) \right. \right. \right.$$
$$\left. \left. \left. - \frac{p(0) - p(t)}{\sigma(t)^2} \right\|_2^2 \right] \right\} \quad (1)$$

where $\epsilon$ is a hyper-parameter representing the minimum noise level. As introduced in [6], to approximate sample the pose candidates, we can solve the following ODE.

$$\frac{dp}{dt} = -\sigma(t)\dot{\sigma}(t)\nabla_p \log p_t(p \mid O) \quad (2)$$

where the score function $\log p_t(p \mid O)$ is approximated by score network $\Phi_\theta(p, t \mid O)$. In practice, the point cloud $O$ is encoded by PointNet++ [31] to extract the feature, and the input pose $p$ and the time variable $t$ are encoded by MLPs.

### C. Symmetric-aware 6D Pose Representation

To handle the Tool's EE symmetry problem, we contribute a symmetric-aware 6D pose representation for jointly learning the symmetry properties and the 6D pose. For the output from the diffusion model, we have four fully connected layers, which output four 3D vectors, in which, $[R_x \mid R_y] \in \mathbb{R}^6$ denote a continuous rotation representation proposed by [32], $S \in \mathbb{R}^3$ denote the symmetry vector to augment the rotation for the symmetry-agnostic task, and with the $T \in \mathbb{R}^3$ denote the translation vector, that our symmetric-aware pose representation is $[R_x \mid R_y \mid S \mid T] \in \mathbb{R}^{12}$. In this way, the model not only knows how to accurately estimate the 6D pose of the target but also be aware of which axis is ambiguous. And it is assistive for the pose post-processing, considering the tool manipulation case, the rotation between the tool and the robot hand can be considered roughly fixed, so we use the orientation of the robot hand to refine the symmetric axis of the estimated pose of each tool's EE. It helps to improve the accuracy and robustness of the estimation when ambiguity exists without using any of the priors. In the experiment of ablation study and qualitative result, we show the effectiveness of our approach.

## V. EXPERIMENTS AND RESULTS

### A. Experiment Setup

Our framework consists of two separate parts, affordance segmentation, and 6D pose estimation. Therefore, the results will be reported individually for better results inspection. Since the pose estimation is our focus, for affordance segmentation, we simply trained it and reported its precision to show that the chosen segmentation algorithm is capable of this task. For evaluating the 6D pose estimation methods, we have carried out comparative experiments, the baselines are detailed in section of ablation study.

### B. Evaluation Metrics

**Metrics of Affordance Segmentation:** As in [9], for evaluating the affordance map segmentation, we report the standard COCO metrics.

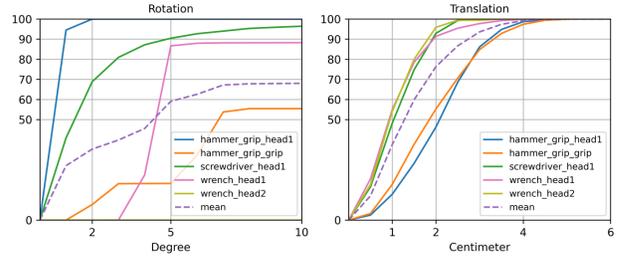

(a) Pose Estimation with Object-centric points cloud

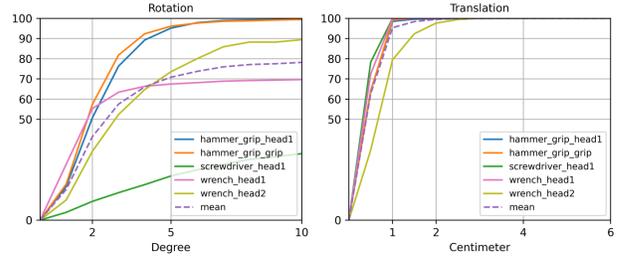

(b) Symmetry-agnostic pose representation

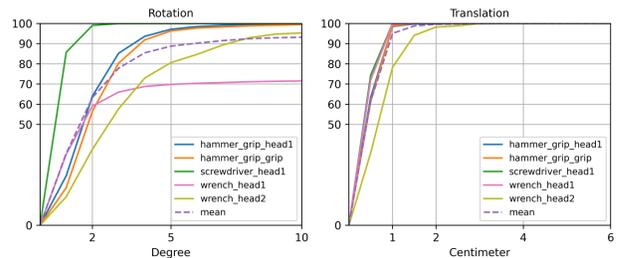

(c) Symmetry-aware pose representation (Ours)

Fig. 5. The ablation study result includes mean average precision of translation and rotation per category. (a) relate to pose estimation with object points cloud and pose prior. (b) relate to pose estimation using symmetry-agnostic pose representation, i.e. $(R_x, R_y, T) \in \mathbb{R}^9$ as in [32]. (c) relate to pose estimation using symmetry-aware pose representation, i.e. $(R_x, R_y, S, T) \in \mathbb{R}^{12}$.

**Metrics of Symmetry-agnostic 6D Pose Estimation:** For evaluating 6D pose estimation, we refer to the methodology outlined in [4] and report the Mean Average Precision ($mAP$) within specified error margins. These margins are defined by rotation error $n°$ and translation error $m$ cm. For instance, $5°2cm$ indicates that the predicted rotation error is less than $5°$ and the translation error is less than 2 cm.

### C. Experimental Results

*1) Affordance Segmentation:* For the affordance map segmentation, we have trained a segmentation module [9] and achieved an overall segmentation Average Precision of 84.16, $AP_{50}$ of 98.17, $AP_{75}$ of 92.27, $AP_S$ of 72.59, $AP_M$ of 92.05, $AP_L$ of 100.

*2) Ablation Study:* We use an ablation study to investigate the effectiveness of our approach, the result is shown in Fig. 5. Specifically, we compare the pose estimation results on the object-centric points cloud and our EE-centric points cloud. Furthermore, we compare the pose estimation result

TABLE I
THE COMPARISON OF POSE ESTIMATION RESULTS

| Category of Tool's EE | Symmetric | 5° 2cm ↑ | | | | 5° 5cm ↑ | | | | 10° 5cm ↑ | | | |
|---|---|---|---|---|---|---|---|---|---|---|---|---|---|
| | | obj_w_pp | ee | ee_w_sp | Our | obj_w_pp | ee | ee_w_sp | Our | obj_w_pp | ee | ee_w_sp | Our |
| Hammer grip | No | 0.067 | 0.961 | 0.961 | **0.967** | 0.182 | 0.961 | 0.961 | **0.967** | 0.554 | **0.995** | **0.995** | 0.994 |
| Hammer head | Yes | 0.463 | 0.951 | 0.973 | **0.976** | **1.0** | 0.951 | 0.974 | 0.976 | **1.0** | 0.992 | 0.993 | **0.997** |
| Screwdriver head | | 0.436 | 0.219 | **1.0** | **1.0** | 0.840 | 0.219 | **1.0** | **1.0** | 0.964 | 0.330 | **1.0** | **1.0** |
| Wrench head | | 0.504 | 0.674 | 0.677 | **0.996** | **0.831** | 0.674 | 0.677 | 0.996 | 0.882 | 0.696 | 0.697 | **0.999** |
| Average | - | 0.368 | 0.701 | 0.903 | **0.984** | 0.713 | 0.701 | 0.903 | **0.984** | 0.850 | 0.753 | 0.921 | **0.998** |

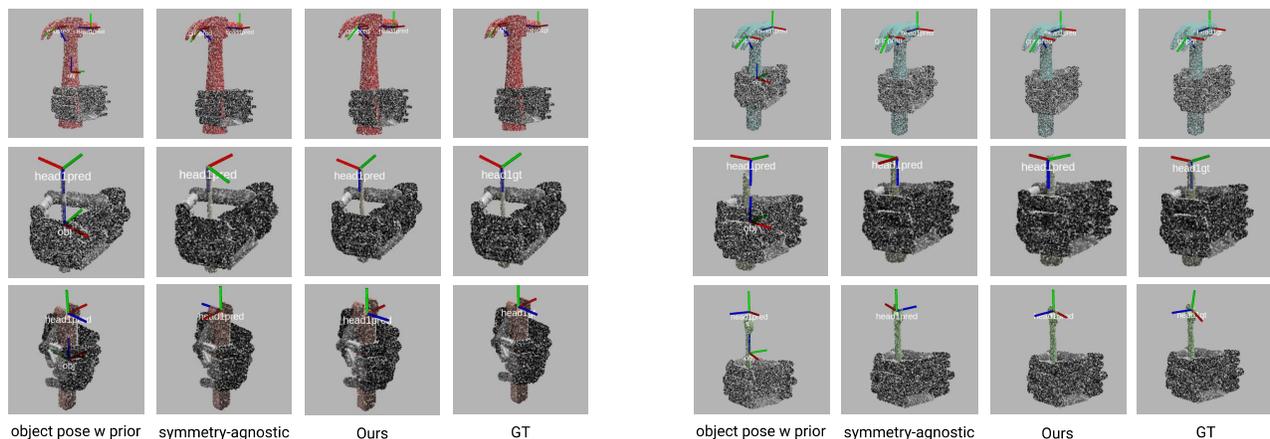

object pose w prior  symmetry-agnostic  Ours  GT      object pose w prior  symmetry-agnostic  Ours  GT

Fig. 6. The qualitative experimental result. The first row shows the result of the hammer, the second is a screwdriver, and the third is a wrench. The left side includes the seen objects, while the right side belongs to the novel objects. Object pose w prior means firstly estimating the pose of the whole object and then applying the category-level pose prior to it. Symmetry-agnostic denotes directly taking the 6D pose output from the model. Ours means we use the leaned symmetry knowledge to tune the symmetric axis based on the model output. GT means the ground truth of the 6D pose.

with and w/o symmetry-aware pose representation. In each comparison, we just change one aspect of our approach. For example, we compare the object-centric points cloud and EE-centric points cloud pose estimation, both using symmetry-aware pose representation.

**Influence of Tool's EE Pose Estimation:** As in Fig. 5 (a) and (c), we can see, that the categorical-level pose estimation based on EE-centric points cloud obviously outperforms the object-centric at the translation and mean of rotation. However, due to the multi-hypothesis issue from the symmetric target, the categorical pose prior shows its benefit for pose disambiguation, and therefore, it is better than the EE-centric pose estimation. Moreover, in Table I, the pose estimation based on the object points cloud($obj\_w\_pp$) is unstable and underperforms other methods at an average level.

**Influence of Symmetry-agnostic 6D Pose Estimation:** Table I shows the performance comparison of our symmetry-aware pose representation with EE-centric point cloud as input, $obj\_w\_pp$ denotes the category-level pose estimation is applied to estimate the 6D pose of the whole object and use the category-level pose prior for extracting the pose of each tool's EE. $ee$ means the pose representation as in [32] is used for estimating the pose from the points cloud of each tool's EE and does not consider the symmetry for metrics calculation. $ee\_w\_sp$ means the priors of symmetry are additionally employed to disambiguate the symmetry result. $our$ means the proposed symmetry-aware pose representation, i.e. the estimated symmetric properties are used for metrics to disambiguate the result. We can see, that our proposed symmetry-aware pose representation outperforms the normal representation and is even slightly better than the priors, which shows the outstanding performance of our method. Moreover, for the per category results in Fig. 5 (b) and (c), we can see our proposed representation has a clear impact on the typical symmetric tool's EE, head of the screwdriver. The reason that there are no obvious improvements in the category head of the hammer is that the affordance map could cover a part of the hammer handle, which makes the points cloud asymmetric.

*3) Qualitative Analysis:* As in Fig. 6, The pose estimation that relies on the whole object's pose prior (*object pose w prior*) is promising for orientation, but less reliable for position estimation. This discrepancy is understandable since the high variance of the shape and scale exists within a category. Furthermore, symmetric objects present a multi-hypothesis challenge, leading to identical geometric configurations in the observed point cloud from different viewpoints. Consequently, the estimated pose (*symmetry − agnostic*) could be flipped or skewed. For this issue, our method jointly learns both the pose and symmetry properties for disambiguation, achieving a result that closely rivals the ground truth.

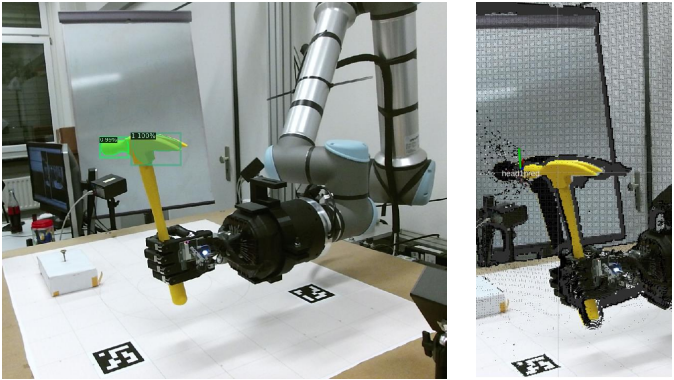

Fig. 7. The real-robot experiment. We use a shadow hand plus UR10 set up to grab a hammer from a human hand-over and use that hammer to drive a nail by estimating and adjusting the hammer's pose after the hand-over. In the picture, the left is the segmentation result, and the right is the pose estimation result of the hammerhead.

*4) Real Robot Experiment:* We also employed our method in real robot experiments for qualitative analysis, as in Fig. 7, specifically for the hammering nails task. In this task, first, a human operator hands over a hammer to the dexterous robot hand; secondly, we employ the Kinect v2 camera to get RGB-D input and then segment the RGB channel to get the semantic mask of hammers EE. For the hammering task, we choose the segmentation of the hammerhead and use that to get the corresponding point cloud from the depth channel. Then, our diffusion model-based pose estimator outputs the estimated pose of the hammerhead for further tool manipulation. The diffusion model-based pose estimator was trained using an Nvidia RTX A6000 graphics card with our TooleeDataset for approximately 13 hours. Leveraging the knowledge of the current pose of a hammerhead, the robot adjusts its position to successfully drive the nail into the block, and the inference time is about 14 fps. Without adjusting the pose of the hammerhead, just use a pre-defined trajectory; the success rate is much lower than with our approach (20% vs. 85% in 20 trials) because it can't guarantee that each handover yields the same in-hand pose of the hammer. Interestingly, even using the synthetic dataset to train our model, the segmentation result and the pose estimation result both still look good. This is because our synthetic dataset is realistic, and the domain gap is relatively small to train the Mask R-CNN. Additionally, our pose estimator, which relies solely on point cloud input, also benefits from a small domain gap within sim and real point clouds, resulting in accurate pose estimation.

## VI. DISCUSSION AND CONCLUSION

This study presents a framework specifically designed for affordance pose estimation in the context of robot tool manipulation leveraging our custom-developed synthetic ToolEE dataset. We study the symmetry problem, proposing a novel symmetry-aware pose representation that enables prior-free 6D pose estimation. Our extensive experimentation validates the rationality and quality of our ToolEE dataset, and our framework and symmetry-aware pose representation outperform baseline methods based on our dataset.

**Limitations and Future Work:** While our experimental results support our hypothesis and validate the proposed method, our work still has some limitations. Increasing the diversity of 3D model assets is crucial to mitigate dataset bias, enhancing the robustness and applicability of our method. In future work, we aim to expand our dataset by incorporating a wider range of tool categories and augmenting it with more real-world data. Furthermore, we plan to conduct extensive experiments with real robots on a variety of tasks. While we have already demonstrated success with hammering nails, future work will also include extracting nails, tightening and loosening nuts with a wrench, using a screwdriver, and exploring additional tasks with tools of new categories.